\documentclass[11pt]{article}

\usepackage[margin=1in]{geometry}
\usepackage{authblk}
\usepackage[skip=0.6\baselineskip, indent=0pt]{parskip}
\usepackage{amsmath}
\usepackage{amssymb}
\usepackage{booktabs}
\usepackage{graphicx}
\usepackage{caption}
\usepackage[hidelinks]{hyperref}

\title{The Query Knows What to Forget: A Second Erase Direction for Linear Attention}
\author[1]{Dhruman Gupta}
\author[1]{Aritra Das}
 \author[1]{Debayan Gupta}
\affil[1]{Truth Audit Labs}
\date{}

\begin{document}
\maketitle

\begin{abstract}
Linear attention keeps a state of fixed size. At long context, many stored
items share this state, and interference between them degrades retrieval.
Gated DeltaNet-2 (GDN-2), like every delta-rule model before it, derives its
erase vector from the key of the current token. However, the interference
in its reads is measured through the query, and the erase step cannot reach it.
We introduce the Query-derived Erase Direction (QED). QED adds a second erase direction derived from the query and
orthogonal to the key. In the fast-weight view, a key-directed delta
edit cannot change the key-orthogonal part of a read. It uses the editable
part to cancel old-state content measured along the query. It also improves retrieval at every
length past the training window, and it about doubles the usable context
length on S-NIAH-1.
\end{abstract}

\section{Introduction}

Linear attention replaces the sequence-length-dependent cache of softmax
attention with a fixed-size recurrent state, giving each layer constant memory
and constant sequence-mixing cost per token during autoregressive decoding
\cite{katharopoulos2020,schlag2021}. This efficiency requires all stored
associations to share the same bounded state. As the context grows, these
associations interfere, and retrieval from recurrent linear models becomes
less reliable \cite{yang2024deltanet,yang2025gdn}. Recent architectures have
therefore made the state update increasingly selective. DeltaNet corrects an
existing association before writing a new value
\cite{schlag2021,yang2024deltanet}, Gated DeltaNet adds adaptive decay
\cite{yang2025gdn}, Kimi Delta Attention makes that decay channel-wise
\cite{kimilinear2025}, and Gated DeltaNet-2 (GDN-2) separates the erase and
write gates \cite{gdn2}.

Despite these changes, the erase vector in each update is still derived from
the current key \cite{schlag2021,yang2024deltanet,yang2025gdn,
kimilinear2025,gdn2}. This design creates a mismatch between reading and
editing the state. The query selects the content returned by a read, whereas a
key-derived vector selects the old content removed by the update. Content
visible to the query can therefore lie outside the part of the state that the
erase step can directly modify.

We address this mismatch by using the query to define a second erase
direction. The Query-derived Erase Direction (QED) augments the GDN-2 erase vector with a
gated term derived from the query and projects this term orthogonal to the key.
In the fast-weight view, the standard delta update cannot edit the
key-orthogonal part of a read. QED uses the added direction to cancel old-state
content that contributes to this part (Section \ref{sec:online}). The change
preserves the rank-one update and the stability condition of GDN-2, with
negligible additional cost.

We train 340M-parameter models on 15B tokens, with two seeds and five arms. The
contributions are:
\begin{enumerate}
  \item We introduce QED, which adds a gated, key-orthogonal erase correction
        derived from the query to the rank-one update of GDN-2
        (Section \ref{sec:method}).
  \item Across two training seeds, QED improves S-NIAH-1 retrieval beyond the
        training length by up to $39$ percentage points and improves FDA by
        $3.4$ and $3.7$ points
        (Section \ref{sec:experiments}).
  \item Our controlled ablations show that using the query to define the added
        erase direction is consistently beneficial. The effects of gating and
        key-orthogonal projection remain inconclusive across runs
        (Section \ref{sec:support}).
\end{enumerate}

\section{Background}
\label{sec:background}

\paragraph{Linear attention and the delta rule.} Linear attention
\cite{katharopoulos2020} keeps a state matrix $S_t \in \mathbb{R}^{K \times V}$
in place of the softmax cache. The base rule is additive:
$S_t = S_{t-1} + k_t v_t^{\top}$, with the read $o_t = S_t^{\top} q_t$. The
additive rule cannot correct a stored value. DeltaNet \cite{schlag2021,
yang2024deltanet} replaces the addition with an error-correcting step:
\begin{equation}
  S_t \;=\; \bigl(I - \beta_t\, k_t k_t^{\top}\bigr)\, S_{t-1}
        \;+\; \beta_t\, k_t v_t^{\top},
  \label{eq:deltanet}
\end{equation}
where $\beta_t \in (0,1)$ is a scalar. The rank-one term erases the old value
at the key, and then the layer writes the new value at the same key.

\paragraph{Gated DeltaNet (GDN).} DeltaNet has no forgetting. Gated DeltaNet
\cite{yang2025gdn} adds a scalar decay $\alpha_t \in (0,1)$ before the delta
step:
\begin{equation}
  S_t \;=\; \alpha_t \bigl(I - \beta_t\, k_t k_t^{\top}\bigr)\, S_{t-1}
        \;+\; \beta_t\, k_t v_t^{\top}.
  \label{eq:gdn}
\end{equation}
The decay gives a Mamba-style forget path, and the delta term gives targeted
correction. One scalar $\alpha_t$ decays every channel of the state at the
same rate.

\paragraph{Kimi Delta Attention (KDA).} KDA \cite{kimilinear2025} makes the
decay channel-wise. A vector $\alpha_t \in (0,1)^{K}$ replaces the scalar:
\begin{equation}
  S_t \;=\; \bigl(I - \beta_t\, k_t k_t^{\top}\bigr)\,
        \mathrm{Diag}(\alpha_t)\, S_{t-1}
        \;+\; \beta_t\, k_t v_t^{\top}.
  \label{eq:kda}
\end{equation}
Each key channel now forgets at its own rate. The single scalar $\beta_t$
still controls two different things: how much old content the erase removes,
and how much new content the write commits.

\paragraph{Gated DeltaNet-2 (GDN-2).} GDN-2 \cite{gdn2} decouples the two
roles of $\beta_t$. A channel-wise erase gate $b_t \in (0,1)^{K}$ controls the
removal, and a channel-wise write gate $w_t \in (0,1)^{V}$ controls the commit:
\begin{align}
S_t &= \bigl(I - k_t\,a_t^{\top}\bigr)\,\mathrm{Diag}\!\left(\exp g_t\right) S_{t-1}
        + k_t\,(w_t \odot v_t)^{\top}, \label{eq:gdn2}\\
o_t &= S_t^{\top} q_t , \nonumber
\end{align}
with the erase vector $a_t = b_t \odot k_t$ and the log-decay
$g_t \in \mathbb{R}^{K}$. GDN-2 reduces to KDA when $b_t$ and $w_t$ collapse
to one scalar, and to GDN when the decay also collapses. The symbols are:
\begin{itemize}
  \item $q_t, k_t \in \mathbb{R}^{K}$ are the query and the key. The layer
        applies L2 normalization to both.
  \item $v_t \in \mathbb{R}^{V}$ is the value, and $w_t$ is its write gate.
  \item $g_t \in \mathbb{R}^{K}$ is the log of the decay. The term
        $\mathrm{Diag}(\exp g_t)$ makes the state forget slowly.
  \item $b_t \in (0,1)^{K}$ is the erase gate.
\end{itemize}
The term $I - k_t a_t^{\top}$ is a rank-one edit. It removes the part of the
state read through the gated key $a_t=b_t\odot k_t$. Then the layer writes the
new value at $k_t$. Thus the erase vector is still built from the current
key, although its channels are gated independently. This shared property of equations (\ref{eq:deltanet}) to
(\ref{eq:gdn2}) is the target of our change. Erase-then-Delta attention
\cite{etd2026} also uses separate erase and write vectors. Our
change differs in that it keeps a single rank-one update, gates the correction
with $b_t$, and constrains the correction to be orthogonal to the key.

\section{Query-derived Erase Direction}
\label{sec:method}

\subsection{Motivation}
\label{sec:motivation}

The state is a sum of decayed outer products, so the read at step $t$ is a
mixture:
\begin{equation}
  o_t \;=\; S_t^{\top} q_t
      \;=\; \sum_{j \le t} \bigl(q_t^{\top} \tilde{k}_j\bigr)\, \tilde{v}_j ,
\end{equation}
where $\tilde{k}_j, \tilde{v}_j$ denote the stored key--value pairs after
decay and past erasures. One term of this sum is the target and every other
term with $q_t^{\top}\tilde{k}_j \neq 0$ is interference. A longer context
stores more pairs in the same fixed-size state, so interference grows with
length, and retrieval degrades.

Note where this interference lives: along $q_t$, because it is by definition
the part of the state that $q_t$ picks up. The erase step cannot remove it.
In every model of Section \ref{sec:background}, the erase vector is $k_t$ or
a gated version of it. QED adds a direction derived from $q_t$: the layer becomes
able to remove content selected by what it reads.

We fix this by adding a second erase direction to the recurrence in \eqref{eq:gdn2}. It
replaces the erase vector $a_t = b_t \odot k_t$ with
\begin{equation}
  a_t \;=\; b_t \odot k_t \;+\; \lambda_h\, P^{\perp}_{k_t}\!\bigl(b_t \odot q_t\bigr),
  \qquad
  P^{\perp}_{k}(x) \;=\; x - k\,\bigl(k^{\top} x\bigr),
  \label{eq:qed}
\end{equation}
where $\lambda_h \ge 0$ is a learned per-head strength. The second term is
derived from the read direction $q_t$, where the motivation above located
the interference. We discuss further details in Appendix \ref{app:lambda}.

The gate $b_t$ scales the correction channel-wise. Without it, the layer
would erase on every read and delete content that later reads still need. The
projection $P^{\perp}_{k_t}$ removes the component along $k_t$. This keeps the
new component of the erase vector key-orthogonal and leaves the nontrivial
eigenvalue of $I-k_ta_t^{\top}$ unchanged. We test each part of
\eqref{eq:qed} with the ablations in Section \ref{sec:setup}, and discuss the
results in Section \ref{sec:support}.

\subsection{The online learning perspective}
\label{sec:online}

The fast-weight view gives a direct way to isolate the effect of QED.
DeltaNet takes a reconstruction step at the current key
\cite{schlag2021}, GDN and KDA first decay the state, GDN-2 makes the
residual channel-selective \cite{gdn2}. Let $\bar{S}_t$ be the state after
decay:
\begin{equation}
  \bar{S}_t = \mathrm{Diag}(\exp g_t) S_{t-1}.
  \label{eq:fastweight-defs}
\end{equation}
GDN-2 reads old content along the erase direction $b_t\odot k_t$ and compares it
with the gated value $w_t\odot v_t$. Its update is therefore
\begin{equation}
  S_t^{\mathrm{GDN2}}
  = \bar{S}_t
    + k_t\!\left[w_t\odot v_t-
      \bar{S}_t^{\top}(b_t\odot k_t)\right]^{\top}.
  \label{eq:gdn2-residual}
\end{equation}
For any probe $x$, subtracting the pre-update response from
\eqref{eq:gdn2-residual} gives
\begin{equation}
  \bigl(S_t^{\mathrm{GDN2}}\bigr)^{\top}x-\bar{S}_t^{\top}x
  = (x^{\top}k_t)
    \left[w_t\odot v_t-\bar{S}_t^{\top}(b_t\odot k_t)\right].
  \label{eq:probe-response}
\end{equation}
Thus a key-directed delta edit cannot change the response to any
$x\perp k_t$. In particular, if
$\rho_t=q_t^{\top}k_t$ and
$q_t^{\perp}=P_{k_t}^{\perp}q_t$, then
$q_t=\rho_t k_t+q_t^{\perp}$ because $\lVert k_t\rVert_2=1$. Multiplying
\eqref{eq:gdn2-residual} by $q_t$ and using this decomposition gives
\begin{align}
  o_t^{\mathrm{GDN2}}
  &= \bar{S}_t^{\top}q_t
     +\rho_t\!\left[w_t\odot v_t-
       \bar{S}_t^{\top}(b_t\odot k_t)\right] \nonumber\\
  &= \rho_t\!\left[w_t\odot v_t+
      \bar{S}_t^{\top}\bigl((\mathbf{1}-b_t)\odot k_t\bigr)\right]
    + \bar{S}_t^{\top}q_t^{\perp}.
  \label{eq:gdn2-read-decomp}
\end{align}
The last term is old-state content read by the part of the query that the
delta edit cannot reach directly.

QED measures this content along the gated direction
$d_t=P_{k_t}^{\perp}(b_t\odot q_t)$ and subtracts it from the residual in
\eqref{eq:gdn2-residual}. Substituting the QED erase vector
$b_t\odot k_t+\lambda_h d_t$ gives the exact identities
\begin{align}
  S_t^{\mathrm{QED}}
    &= S_t^{\mathrm{GDN2}}
       -\lambda_h k_t\bigl(\bar{S}_t^{\top}d_t\bigr)^{\top},
       \label{eq:qed-state-difference}\\
  o_t^{\mathrm{QED}}
    &= o_t^{\mathrm{GDN2}}
       -\lambda_h\rho_t\bar{S}_t^{\top}d_t.
       \label{eq:qed-read-difference}
\end{align}
Equation \eqref{eq:qed-read-difference} is the full effect of the mechanism on
the current read. The correction samples the old state in a gated,
direction derived from the query, writes the negative sample along the key, and reaches
the query through the key--query overlap $\rho_t$.


\section{Experimental setup}
\label{sec:setup}

\paragraph{Baseline model.} We use GDN-2, equation (\ref{eq:gdn2}), in a
24-layer decoder. Each block has one GDN-2 layer and one SwiGLU layer with RMSNorm.

\subsection{Ablation arms}

Table \ref{tab:arms} shows the arms. Each arm removes one part of the mechanism.
The arms let us find which part does the work.

\begin{table}[h]
\centering
\caption{The arms. \emph{Gate} is the $b_t$ gate on the correction.
\emph{Project} is $P^{\perp}_{k_t}$. \emph{Direction} is the source of the
correction in equation (\ref{eq:qed}).}
\label{tab:arms}
\begin{tabular}{llll}
\toprule
Arm & Gate & Project & Direction \\
\midrule
GDN-2 & --- & --- & no correction \\
QED-nogate-noproj & no  & no  & query \\
QED-nogate & no  & yes & query \\
QED (full) & yes & yes & query \\
QED-rand (control) & yes & yes & fixed random unit vector for each head \\
\bottomrule
\end{tabular}
\end{table}

The QED-rand arm is the direction control. It keeps the gate and the projection,
but it replaces the query with a fixed random direction. If the query-derived direction is
necessary, then QED-rand must fall back to the baseline. The QED-nogate arm is the gate
control. It keeps the query direction and the projection, but it removes the
gate.

\subsection{Training and evaluation}

\paragraph{Training.} All models have 340M parameters: 24 layers, hidden size 1024,
6 heads, head dimension 128, tied embeddings. The data is FineWeb-Edu, tokenized
with the Llama-2 tokenizer. The train length is 2048 tokens. The optimizer is
AdamW at learning rate $3 \times 10^{-4}$, with a WSD schedule and a global batch
of 524{,}288 tokens. Each main run trains on 15B tokens, and the decay starts at
13.5B tokens. The precision is BF16 with FP32 master weights. All five arms ran with seed 0
and with seed 1.

\paragraph{Evaluations.} We use RULER S-NIAH-1, S-NIAH-2 and S-NIAH-3 with 500
samples for each length, from 1K to 32K tokens. We use the Based recall tasks:
SWDE, FDA and SQuAD-completion.

\paragraph{Inference-time strength test.} For each final QED checkpoint,
we keep all weights fixed and replace the learned strength with
$\lambda_h^{(s)}=s\lambda_h$ for
$s\in\{-1,0,0.5,1,1.5,2\}$. The scale is applied at every token during both
prefill and decoding, so it changes the full recurrent trajectory rather than
only the final output. Every setting uses the same deterministic S-NIAH-1
examples, and we compare it with the trained setting $s=1$ per example using
the exact McNemar test. This test measures the contribution of
the QED recurrence for fixed trained weights. We note that it is not a training-time sweep
over $\lambda_{\max}$. Appendix \ref{app:lambda-sweep} gives the results and
further limitations.

\section{Results}
\label{sec:experiments}

\subsection{Language modelling quality}

Table \ref{tab:val} shows that QED does not change the validation loss. This is
the expected result: the mechanism targets retrieval, and not perplexity.

\begin{table}[h]
\centering
\caption{Final validation loss at 15B tokens (nats). The full spread across
all ten runs is 0.0034 nats. Within each seed, the spread is at most 0.0013
nats. The best value for each seed is bold.}
\label{tab:val}
\begin{tabular}{lccccc}
\toprule
Seed & GDN-2 & QED (full) & QED-rand & QED-nogate & QED-nogate-noproj \\
\midrule
0 & 2.3714 & 2.3719 & \textbf{2.3712} & 2.3718 & 2.3724 \\
1 & 2.3746 & 2.3735 & 2.3735 & \textbf{2.3733} & 2.3741 \\
\bottomrule
\end{tabular}
\end{table}

\subsection{Long-context retrieval}

\begin{figure}[t]
\centering
\includegraphics[width=\textwidth]{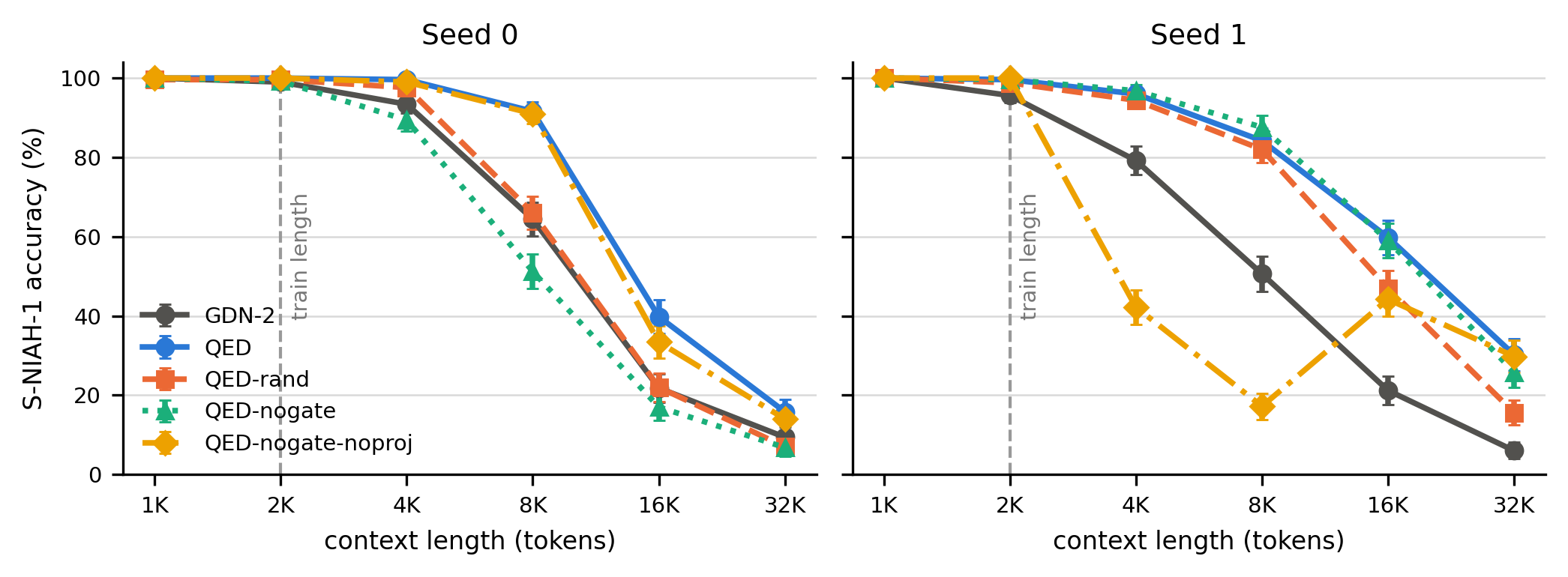}
\caption{S-NIAH-1 accuracy against context length, for each arm and each
seed. The dashed line marks the 2048-token training length. Error bars are
95\% binomial intervals. QED is above GDN-2 at every length past the training
length, in both seeds. The ablation arms do not agree between the seeds.}
\label{fig:niah1}
\end{figure}

Table \ref{tab:niah1} and Figure \ref{fig:niah1} show S-NIAH-1. The training
length is 2048 tokens. Thus all lengths from 4K to 32K are extrapolation. The
full QED arm increases the accuracy at every length above the training length,
in both seeds.

\begin{table}[h]
\centering
\caption{S-NIAH-1 accuracy (\%) at each context length, with the recall
tasks FDA, SQuAD-completion and SWDE (accuracy, \%). All rows are 15B
tokens. The best value for each seed and each column is bold.}
\label{tab:niah1}
\small
\setlength{\tabcolsep}{3.5pt}
\begin{tabular}{llrrrrrrrrr}
\toprule
Seed & Arm & 1K & 2K & 4K & 8K & 16K & 32K & FDA & SQuAD & SWDE \\
\midrule
0 & GDN-2      & \textbf{100.0} & 99.0 & 93.4 & 64.4 & 21.8 & 9.4 & 4.45 & 29.39 & 19.08 \\
0 & QED  & \textbf{100.0} & \textbf{100.0} & \textbf{99.6} & \textbf{91.6} & \textbf{39.8} & \textbf{15.8} & \textbf{7.80} & 29.29 & \textbf{21.51} \\
0 & QED-rand  & 99.6 & 99.6 & 97.6 & 66.0 & 22.0 & 7.0 & 6.44 & 29.22 & 19.89 \\
0 & QED-nogate & \textbf{100.0} & 99.2 & 89.4 & 51.2 & 17.0 & 6.8 & 7.35 & 28.82 & 19.80 \\
0 & QED-nogate-noproj & \textbf{100.0} & \textbf{100.0} & 99.0 & 91.0 & 33.4 & 14.0 & 4.54 & \textbf{29.93} & 19.98 \\
\midrule
1 & GDN-2      & \textbf{100.0} & 95.6 & 79.2 & 50.6 & 21.2 & 6.0 & 4.26 & \textbf{32.37} & \textbf{19.62} \\
1 & QED  & \textbf{100.0} & 99.6 & 96.0 & 84.2 & \textbf{59.8} & \textbf{30.2} & \textbf{7.99} & 32.21 & \textbf{19.62} \\
1 & QED-rand  & \textbf{100.0} & 98.8 & 94.4 & 82.0 & 47.0 & 15.6 & 5.17 & 30.80 & 18.36 \\
1 & QED-nogate & \textbf{100.0} & 99.6 & \textbf{96.8} & \textbf{87.6} & 59.0 & 25.8 & 6.35 & 29.69 & 17.73 \\
1 & QED-nogate-noproj & \textbf{100.0} & \textbf{100.0} & 42.2 & 17.2 & 44.2 & 29.8 & 5.81 & 29.36 & 18.00 \\
\bottomrule
\end{tabular}
\end{table}

\subsection{Inference-time strength}
\label{sec:lambda-results}

Changing the correction strength gives different results for the two QED
checkpoints. For one checkpoint, accuracy increases at every context length as
the strength increases. At 16K, accuracy is 44.8\% when the correction is
reversed, 55.0\% when it is removed, 65.6\% at the trained strength, and 73.0\%
when the strength is doubled. For the other checkpoint, accuracy stays within
2.4 points of the trained setting and does not move in one direction. Thus a
fixed checkpoint can use the correction, but this result is not consistent
across checkpoints. Appendix \ref{app:lambda-sweep} gives the complete results
and limitations.

\begin{figure}[t]
\centering
\includegraphics[width=\textwidth]{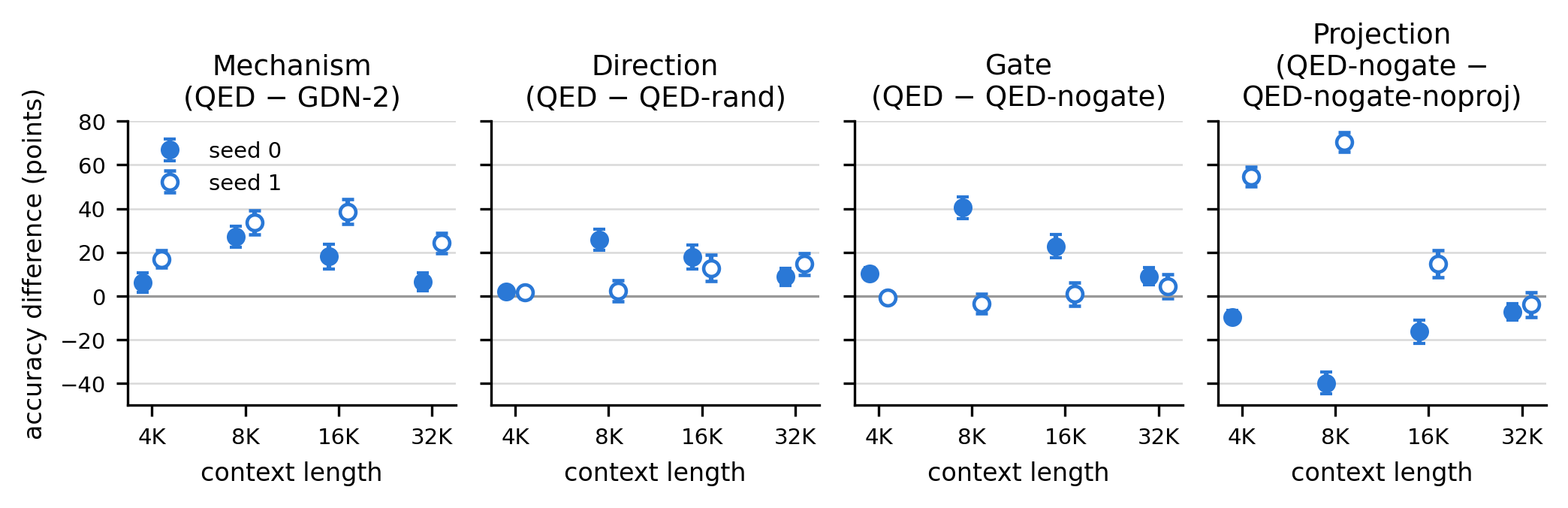}
\caption{The four contrasts that the arms isolate, both seeds, with 95\%
intervals. The mechanism and the direction keep one sign at both seeds. The
gate and the projection reverse between the seeds.}
\label{fig:contrasts}
\end{figure}

\subsection{Document recall}

\begin{figure}[t]
\centering
\includegraphics[width=0.6\linewidth]{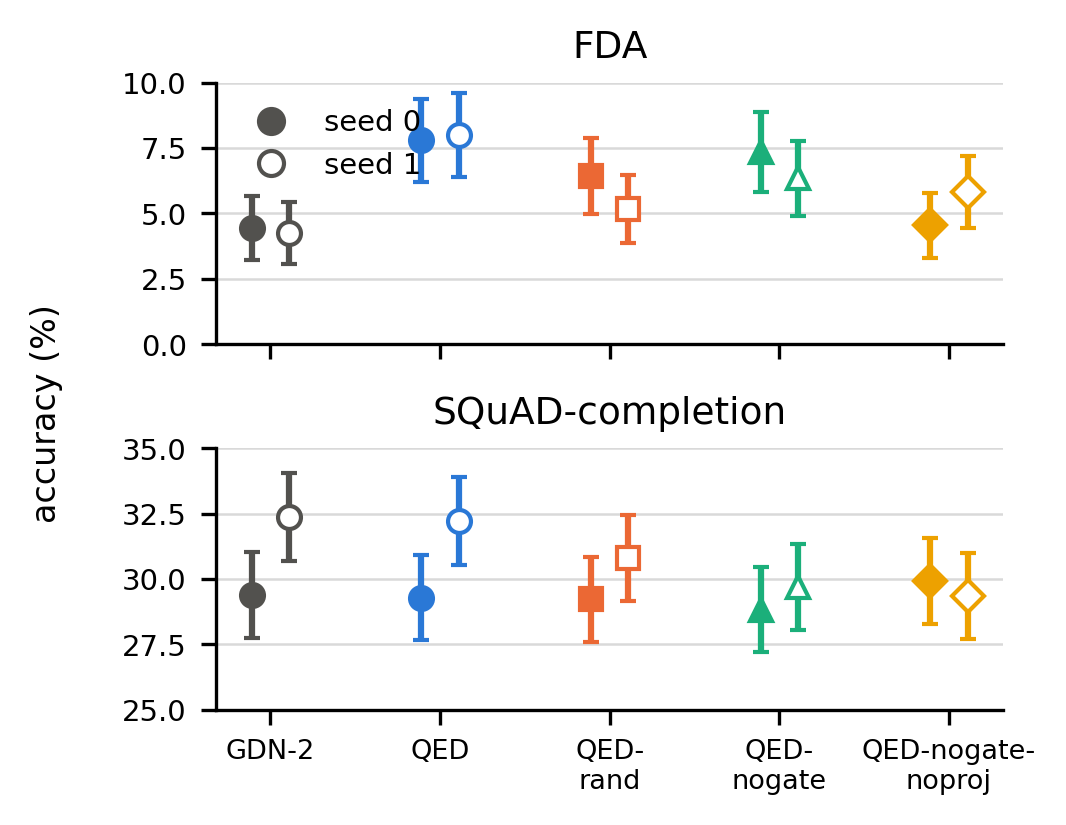}
\caption{FDA and its comprehension control, with 95\% intervals. Every QED
arm is above GDN-2 on FDA, in both seeds. SQuAD-completion does not separate
the arms.}
\label{fig:recall}
\end{figure}

Table \ref{tab:niah1} and Figure \ref{fig:recall} also show the recall
tasks. The recall documents are short. Thus these tasks measure recall, and
not extrapolation. Every arm with the projection increases FDA above the
baseline, in both seeds.
SQuAD-completion does not change. SWDE gives a
small gain at seed 0 and no gain at seed 1, so it is a null result.

FDA is sensitive to training progress. All reported comparisons therefore use
checkpoints trained on the same 15B-token schedule.

\section{Discussion}
\label{sec:support}

The results indicate that QED substantially improves long-context retrieval in
GDN-2. Across both training seeds, QED improves S-NIAH-1 accuracy at every
evaluated length beyond the 2048-token training window and improves FDA
recall. These gains occur without a measurable change in validation loss or
SQuAD-completion, suggesting that QED improves access to the compressed
recurrent memory rather than language modelling quality in general. The
improvement is also practically meaningful: at comparable retrieval accuracy,
QED extends the useful context from approximately 8K to 16K tokens.

The ablations most consistently support using the query to define the added
erase direction. QED outperforms the random-direction control at every extrapolation
length in both seeds. In contrast, whether the erase gate and key-orthogonal
projection provide additional benefits remains undecided because their effects
are not consistent across runs. Consequently, the experiments support the
complete QED update and its query-derived direction, but do not establish either
sub-component as individually necessary.

To further understand why the effect of the projection is inconclusive, we
analysed the nontrivial eigenvalue of the state transition. Without projection,
this eigenvalue
$e=1-a_t^{\top}k_t$ can in principle lie in $[-0.25,1.25]$, whereas projection
keeps the additional correction orthogonal to $k_t$. In 1.18M token-head
samples from the two unprojected models, however, the observed eigenvalue
always remained in $[-0.011,0.985]$. Counterfactually projecting the learned
correction changed this eigenvalue by less than $10^{-3}$, indicating that the
unprojected correction already had negligible alignment with $k_t$. At this
scale, training therefore avoids the expansive regime that the projection is
designed to exclude. The projection
remains a safeguard, but its constraint is effectively inactive for the
learned states measured here. A larger $\lambda_{\max}$, a larger model, or a
weaker erase gate could reduce this margin and make the constraint relevant.

\subsection{Inference-time strength test}

Changing the correction strength at inference time gave different results
across training runs. Accuracy increased with strength for one run. Changing
the strength had little effect on the other run. Thus the test does not show
how much QED-trained models rely on the correction. It supports a training-time
test of different $\lambda_{\max}$ values. It does not show that the value used
here constrained training. Appendix \ref{app:lambda-sweep} reports the full
results.

Ablation effects also vary substantially across training runs. We therefore
draw component-level conclusions only from effects that are consistent across
the matched 15B runs. Additional training runs and a training-time sweep over
$\lambda_{\max}$ are needed to establish the remaining effects.

\section{Conclusion and Further Work}

We introduced the Query-derived Erase Direction (QED), which augments the erase vector with a
correction derived from the query while preserving the rank-one recurrent update. In
340M-parameter models trained on matched 15B-token schedules, QED consistently
improves long-context retrieval beyond the training length and improves FDA
recall, without a measurable change in validation loss or SQuAD-completion.
The ablations most consistently support using the query to define the added
erase direction. The benefits of the erase gate and key-orthogonal projection
remain undecided.

Further work should evaluate QED across more training runs, model scales,
context lengths, and retrieval tasks. A training-time sweep over
$\lambda_{\max}$ is needed to determine the useful correction strength, and
larger-scale experiments should test when the stability constraint imposed by
the projection becomes active.

\clearpage
\appendix
\section*{Appendix}

\section{The strength parameter}
\label{app:lambda}

The strength $\lambda_h$ is one learned scalar for each layer and head. A
sigmoid gate sets it:
\begin{equation}
  \lambda_h \;=\; \lambda_{\max}\,\operatorname{sigmoid}(\theta_h),
  \qquad \lambda_{\max} = 0.25 .
\end{equation}
The bound $\lambda_{\max}$ caps the correction. Each $\theta_h$ is
initialized at $-2$, so training starts at
$\lambda_h = \lambda_{\max}\,\operatorname{sigmoid}(-2) \approx 0.03$: the correction is
active, but small. $\theta$ trains with a learning-rate multiplier of $10$.
At the default rate, $\theta$ moves very slowly: in 1B-token test runs, the
mean of $\lambda$ moved less than $1\%$ from its initial value.

At $\lambda = 0$, equation (\ref{eq:qed}) reduces to equation
(\ref{eq:gdn2}). Since QED only edits the erase vector, the QED kernel
keeps the structure of the GDN-2 kernel. The extra work is one dot product
and one subtraction for each token and head.

\subsection{Inference-time strength test}
\label{app:lambda-sweep}

We test whether the trained models actively use the QED correction at
inference. For each final QED checkpoint, we replace every learned strength by
\begin{equation}
  \lambda_h^{(s)} = s\lambda_h,
  \qquad s\in\{-1,0,0.5,1,1.5,2\},
  \label{eq:lambda-sweep}
\end{equation}
with all model weights fixed. Because $\lambda_h$ is linear in
$\lambda_{\max}$, this is equivalent to changing the effective
$\lambda_{\max}$ after training while holding $\theta_h$ fixed. The change is
applied during both prefill and decoding. Each scale uses the same 500
deterministic S-NIAH-1 examples at each length. We compare it with $s=1$ on
the same examples using the exact McNemar test.

The two checkpoints do not give the same result (Table
\ref{tab:lambda-sweep}). For one checkpoint, accuracy increases at every
context length when the correction strength increases. At 16K, accuracy is
44.8\% with the correction reversed and 55.0\% with it removed. Accuracy is
65.6\% at the trained strength and 73.0\% when the strength is doubled. These
changes agree with the cancellation mechanism. For the other checkpoint,
changing the strength changes accuracy by no more than 2.4 points. Accuracy
does not consistently increase or decrease. FDA also does not change in a
consistent way. Thus
a trained checkpoint can use the correction, but the result is not consistent
across checkpoints.

\begin{table}[h]
\centering
\caption{Inference-time strength sweep for both QED checkpoints. The weights
are fixed, and $s=1$ is the value used during training. Entries other than
$\overline{\lambda}_h$ are accuracies (\%).}
\label{tab:lambda-sweep}
\setlength{\tabcolsep}{10pt}
\begin{tabular}{llrrrrrr}
\toprule
Seed & $s$ & $\overline{\lambda}_h$ & 4K & 8K & 16K & 32K & FDA \\
\midrule
0 & $-1$  & $-0.101$ & 100.0 & 91.2 & 39.8 & 17.8 & 6.99 \\
0 & $0$   & $0$      & 100.0 & 90.0 & 41.4 & 17.6 & 7.53 \\
0 & $0.5$ & $0.051$  & 100.0 & 89.6 & 42.0 & 18.4 & 7.53 \\
0 & $1$   & $0.101$  & 100.0 & 89.2 & 41.8 & 18.0 & 7.80 \\
0 & $1.5$ & $0.152$  & 100.0 & 90.4 & 41.6 & 17.0 & 7.80 \\
0 & $2$   & $0.203$  & 99.8  & 91.4 & 43.6 & 15.6 & 7.62 \\
\midrule
1 & $-1$  & $-0.120$ & 91.6 & 78.0 & 44.8 & 17.0 & 7.99 \\
1 & $0$   & $0$      & 93.2 & 79.8 & 55.0 & 23.2 & 7.53 \\
1 & $0.5$ & $0.060$  & 94.8 & 82.0 & 61.0 & 26.4 & 7.80 \\
1 & $1$   & $0.120$  & 97.2 & 83.6 & 65.6 & 32.6 & 7.71 \\
1 & $1.5$ & $0.180$  & 98.0 & 86.4 & 69.8 & 35.0 & 7.89 \\
1 & $2$   & $0.241$  & \textbf{99.0} & \textbf{89.0} & \textbf{73.0} &
          \textbf{35.4} & 7.44 \\
\bottomrule
\end{tabular}
\end{table}

This test changes the full recurrence, not only the final read. For a
fixed incoming state, equation \eqref{eq:qed-read-difference} is linear in $s$:
\begin{equation}
  o_t^{(s)}-o_t^{(0)}
  =-s\lambda_h\rho_t\bar{S}_t^{\top}d_t.
\end{equation}
Every earlier update in the sequence also uses $s$. Thus the incoming state
$\bar{S}_t$ depends on $s$. The test measures the accumulated effect of the QED
recurrence, not only the final read. Since $d_t\perp k_t$, changing $s$ does
not change the nontrivial eigenvalue of the projected transition. Therefore,
eigenvalue instability does not cause the accuracy changes.

The gain at $s=2$ for one checkpoint supports a test with a stronger
correction during training. It does not show that
$\lambda_{\max}=0.25$ was a binding constraint, because changing the bound
during training can alter both $\theta_h$ and the remaining model weights. A
training-time $\lambda_{\max}$ sweep is needed to answer that question.

\bibliographystyle{plain}
\bibliography{references}

\end{document}